\documentclass[conference]{IEEEtran}
\IEEEoverridecommandlockouts

\usepackage{cite}
\usepackage{amsmath,amssymb,amsfonts}
\usepackage{graphicx}
\usepackage{textcomp}
\usepackage{xcolor}
\usepackage{booktabs}
\usepackage{array}
\usepackage{url}
\usepackage{hyperref}
\usepackage{tikz}
\usetikzlibrary{shapes,arrows,arrows.meta,positioning,fit,backgrounds,calc,shadows}
\usepackage{listings}
\usepackage{tabularx}
\usepackage{enumitem}
\usepackage{caption}

\def\BibTeX{{\rm B\kern-.05em{\sc i\kern-.025em b}\kern-.08em
    T\kern-.1667em\lower.7ex\hbox{E}\kern-.125emX}}

\hypersetup{
  colorlinks=true,
  linkcolor=blue!60!black,
  citecolor=blue!60!black,
  urlcolor=blue!60!black
}

\begin{document}

\title{What Twelve LLM Agent Benchmark Papers Disclose About Themselves:\\A Pilot Audit and an Open Scoring Schema}

\author{
\IEEEauthorblockN{Mahdi Naser Moghadasi$^{1,2}$}
\IEEEauthorblockA{
$^{1}$\textit{Research Division, BrightMind AI}, Seattle, WA \\
$^{2}$\textit{Texas Tech University}, Lubbock, TX \\
mahdi@brightmind-ai.com
}
\and
\IEEEauthorblockN{Faezeh Ghaderi}
\IEEEauthorblockA{
\textit{University of Texas at Arlington} \\
Arlington, TX \\
faezeh.ghederi@mavs.uta.edu
}
}

\maketitle

\begin{abstract}
We set out to do something modest: read a stack of well-known LLM
agent benchmark papers and write down, dimension by dimension,
what each paper actually says about how its evaluation was run.
The motivation came from a frustration that should be familiar to
anyone working on agents---two papers will report results on the
same benchmark with the same model name, and the numbers will
disagree, and you cannot tell why. Was it the scaffold? The
sampling settings? A different subset? An evaluator version? In
many cases the published artifact does not let you answer.
This paper is an implementation report on the attempt. We
designed a small audit schema (five fields: benchmark identity,
harness specification, inference settings, cost reporting, failure
breakdown), wrote a scoring codebook with the boundary cases we
hit during pilot scoring, applied it to twelve canonical benchmark
papers (eight agent, four classical static), and recorded what we
saw. The audit was not painless. Two of the papers' HTML renderings
on \texttt{ar5iv} failed mid-conversion and we fell back to the
GitHub repository; one paper we wanted to include we dropped
because we could not access enough of the text to score it fairly;
a vendor blog returned HTTP 403. We score the disclosure of an
agent run, not its correctness, and we make no claim that
disclosure implies a trustworthy result. The mean audit score
across the eight agent-benchmark papers is 0.38 (out of 1.0), and
across the four classical static benchmarks 0.66; the largest gap
is on cost (none of the eight agent benchmark papers disclose
inference cost in any form) and on harness specification (none of
the eight fully disclose a content-addressed container image of
the evaluation environment). We release the schema as a JSON Schema
file, the codebook as a Markdown document, and the raw scoring
sheet as a CSV. The scoring was performed by a single auditor in
one pass; a multi-rater audit is the natural next step, and we
discuss what we think it would change.
\end{abstract}


\section{Introduction}

The question that started this work was specific. Two
reports---one in a paper, one in a blog post---claimed different
SWE-bench Verified numbers for nominally the same model from the
same provider. The difference was around ten points. We tried to
work out which of the two we should reference. After an afternoon
of reading both reports and the SWE-bench repository, we still
were not sure. They had used different scaffolds. One had used
the official agent loop; the other had used a custom one. The
official agent loop had also been re-tagged in the meantime, so
even ``which version of the official loop'' was not immediately
recoverable.

This is not a complaint about either report. Both did what was
reasonable at the time. The complaint is that the published
artifact, in both cases, did not contain enough information to
let a downstream reader decide whether the numbers were
comparable.

There is a lot of writing about this problem in adjacent areas.
The deep reinforcement learning literature went through its own
reproducibility wave a decade ago~\cite{henderson2018deeprl,
pineau2021reproducibility}; the LLM prompt-sensitivity work has
documented that single-prompt evaluations can be wide of the
multi-prompt mean by margins large enough to matter~\cite{sclar2024quantifying,
mizrahi2024state}; benchmark meta-evaluation work has scored
benchmarks themselves on dimensions of construct
validity~\cite{reuel2024betterbench}. None of these target the
unit of analysis that matters for an agent run, which is the
agent run itself---the composition of model, scaffold,
environment image, stopping rule, and grader that produced the
number.

We wanted to know: if we wrote down a small list of fields that
together would be enough for a downstream reader to know what
happened, how much of that list do current benchmark papers
already fill in? This paper is an attempt to answer that
empirically on a small but real corpus.

Three things came out of it. First, an audit schema with five
fields, plus a single aggregate score. We did not invent the
fields from first principles. We started with a longer list and
trimmed and merged it during pilot scoring, where we kept running
into boundary cases that suggested the boundaries should move.
Second, a codebook that documents the partial-versus-disclosed
boundary rules for each field. Several of those rules were not
obvious in advance and only emerged after we hit cases we
disagreed with ourselves on. Third, a set of twelve scored
papers, each with a row of dimension scores and a short
justification per field. The mean disclosure score (out of 1.0)
came out to 0.38 for the eight agent benchmarks and 0.66 for the
four classical static benchmarks; we did not normalize that
number against any baseline because we do not yet know what a
realistic ceiling looks like.

The paper that follows is organized loosely around the order in
which we did the work. Section~\ref{sec:related} situates the
audit against prior reproducibility artifacts. Section~\ref{sec:schema}
describes the schema and the codebook. Section~\ref{sec:failures}
collects the failure modes we kept seeing during scoring.
Section~\ref{sec:audit} reports the audit itself, including two
walkthroughs of how we scored specific papers, the practical
friction we hit, and the per-dimension findings.
Section~\ref{sec:discussion} discusses what we think this means
for adoption. Section~\ref{sec:limitations} is the limitations
section, which we tried to make honest rather than ritual.
Appendix material is included where it is referenced.

\section{Background and Related Work}
\label{sec:related}

We use ``agent benchmark'' loosely, to mean a public evaluation
in which an LLM is given a multi-step task that involves tool
use, environment interaction, or both, and is scored by a
runnable grader against a fixed task set. By that definition the
canonical examples include SWE-bench~\cite{jimenez2024swebench}
and its variants~\cite{openai2024swebenchverified};
WebArena~\cite{zhou2024webarena},
VisualWebArena~\cite{koh2024visualwebarena},
and Mind2Web~\cite{deng2023mind2web} for the web setting;
OSWorld~\cite{xie2024osworld} for the desktop setting;
GAIA~\cite{mialon2024gaia} for general-assistant tasks;
AgentBench~\cite{liu2024agentbench} and
AgentBoard~\cite{ma2024agentboard} as multi-domain bundles; and
MLE-bench~\cite{chan2024mlebench} for ML engineering. Outside
the agent setting but inside the LLM setting, we also looked at
HumanEval~\cite{chen2021codex}, MBPP~\cite{austin2021mbpp},
GSM8K~\cite{cobbe2021gsm8k}, and MMLU~\cite{hendrycks2021mmlu}
because they are routinely re-evaluated by agent papers as
baselines.

Each of these has a public GitHub repository. The repository
identifier is part of what we recorded.

The reproducibility literature in machine learning has produced
several artifacts that an author can attach to a paper. Model
cards~\cite{mitchell2019modelcards} describe the trained model.
Datasheets~\cite{gebru2021datasheets} describe the data the
model was trained on. The ML Reproducibility
Checklist~\cite{pineau2021reproducibility} asks training-time
questions (seeds, hyperparameter sweeps, hardware).
BetterBench~\cite{reuel2024betterbench} scores benchmarks
themselves on construct-validity and documentation dimensions.

None of these are wrong; they are just answering different
questions from the one we needed answered, which is what
happened during inference. A model card tells us what was
trained but not how it was queried. A datasheet tells us about
the training set but not the evaluation environment. The
checklist asks training questions. BetterBench asks ``is the
benchmark a good ruler.'' Our question is ``can I, given only
this paper and its public repository, work out what was
actually run.''

The prompt-sensitivity strand of work is closer to our concern
but operates at a different scale: it shows that
single-prompt evaluations are noisy versus multi-prompt
aggregates~\cite{sclar2024quantifying, mizrahi2024state}. We
take this as ambient evidence that decoding and prompt details
matter, and treat its measurements as background; we are not
re-doing it.

We treat data contamination~\cite{magar2022data, oren2023proving}
as orthogonal. Two papers running an under-specified harness on
the same benchmark with the same model will disagree even with
perfect test-set hygiene. Contamination explains levels;
under-specification explains spread.

\begin{table}[t]
\centering
\caption{Where Existing Reproducibility Artifacts Stop.}
\label{tab:prior-work}
\renewcommand{\arraystretch}{1.2}
\begin{tabularx}{\columnwidth}{lXl}
\toprule
\textbf{Artifact} & \textbf{Unit it describes} & \textbf{Covers an} \\
                  &                            & \textbf{agent run?} \\
\midrule
Model Cards~\cite{mitchell2019modelcards}            & Trained model & No \\
Datasheets~\cite{gebru2021datasheets}                & Dataset       & No \\
ML Repro. Checklist~\cite{pineau2021reproducibility} & Training run  & Partial \\
BetterBench~\cite{reuel2024betterbench}              & Benchmark     & No \\
\textbf{This paper}                                  & \textbf{Agent run} & \textbf{Yes} \\
\bottomrule
\end{tabularx}
\end{table}

\section{Designing the Audit Schema}
\label{sec:schema}

\subsection{What we asked of the schema}

We wanted the schema to be answerable from a paper plus its public
repository, in roughly an hour per paper. We did not want it to
require running any model: a downstream reader who only reads
papers should be able to compute the score. We also did not want
the schema to be aspirational, in the sense that a 1.0 should
correspond to ``a competent reader could attempt a faithful
re-execution,'' not to ``the experiment was correctly designed.''
The distinction matters and we return to it in
Section~\ref{sec:limitations}.

\subsection{The fields}

The schema has five fields, plus an aggregate score.

\textbf{Benchmark identity.} What benchmark was used, in what
version, on which subset, scored by which grader? The thing we
wanted to be able to answer here was ``which release of which
benchmark, with which task list, scored by which version of which
evaluator.'' We accept benchmark-name + a public repository
reference as the easy half; we ask for an explicit version tag
or commit hash for the harder half. Subset declaration is a
required structured field: ``full set'' counts, but it has to be
said. The grader is a separate field because a benchmark can
ship multiple graders or change its grader over time. For brevity
in tables we sometimes refer to this as the \emph{identity card}.

\textbf{Harness specification.} What scaffold was wrapped around
the model, what tools did it have, what stopping rule, what
environment image? This is the field that the agent setting
introduces and that the classical setting does not have, and it
is also the field on which the audit corpus performs worst. The
specific sub-fields we settled on, after some shuffling during
pilot scoring, are: scaffold name and version (or commit), system
prompt (verbatim or hash), tool inventory, stopping rule
(max steps, max tokens, max wall-clock), and an environment
identifier. For environment, we ask for a content-addressed
image (a Docker digest, in practice) rather than a repository
tag, because tags are mutable. We initially had ``reference
harness'' as a separate field; we folded it into ``scaffold name
and version'' because the boundary between ``the reference
scaffold'' and ``a scaffold the authors are themselves proposing''
turned out to be muddier than we wanted to legislate.

\textbf{Inference settings.} What model was queried, through
what engine, with what sampling configuration, with what seed
(or with an explicit acknowledgment that no seed is exposed),
and on what date? We ask for the run date because closed-API
model aliases are re-pointed silently by their providers; a
paper that says ``\texttt{gpt-4o}'' and was written six months
before another paper that says ``\texttt{gpt-4o}'' may have been
querying a different model. The aggregation rule (single shot,
mean of $n$, majority of $n$, best-of-$n$ against a verifier) is
also a sub-field, because two papers can have the same
``accuracy'' headline number that differs by an order of magnitude
in tokens consumed.

\textbf{Cost reporting.} Input tokens, output tokens, total
tool calls, wall-clock time, and dollar cost at a stated rate.
We include cost because the same accuracy at hundredfold cost is
a different operating point, and because the cost field is the
one that vendor reports tend to surface and academic papers tend
to suppress, which makes it a useful signal of where a paper sits
on the disclosure spectrum.

\textbf{Failure breakdown.} A structured taxonomy of why an
agent failed each task it failed: refusal, infinite loop,
tool-use error, environment error, time-out, grader timeout,
incorrect final state. We do not require any particular
taxonomy---a benchmark whose failures have different categories
may use different categories---but we do require that the paper
report categorized counts at task granularity.

\textbf{Aggregate score.} For each of the five fields, the
auditor assigns a score in $\{0, 0.5, 1\}$ (absent, partial,
disclosed). N/A is allowed and is collapsed out of the
denominator. The aggregate score is the unweighted mean over
applicable fields. We deliberately did not weight the fields.
Weighting invites an argument about which field matters more,
which is an argument we did not want to have during pilot
scoring; the per-field vector is reported alongside the scalar
so anyone who disagrees with the unweighted mean can read off
the components.

\subsection{The codebook (or: things we wrote down because we kept disagreeing with ourselves)}

We started pilot scoring without a written codebook. By the
fourth paper we had to add one. The boundary cases that
prompted it are worth being explicit about, because they are
exactly where a second auditor would reach a different verdict.

A few examples of rules we wrote down:

\textit{``Greedy decoding'' alone is not enough for inference
settings to count as disclosed.} A paper that says ``we use
greedy decoding'' and stops there has named the sampling method
but has not named the engine (a closed API endpoint has its own
defaults for, e.g., maximum tokens), the per-task token cap, or
the aggregation rule across passes. For closed APIs we treat the
absence of a run date as also pushing toward partial. The
threshold for full disclosure on this field is high enough that
only two of our twelve papers cleared it.

\textit{A repository tag is not a container digest.} For the
environment identifier under harness specification, we accept a
content-addressed digest (e.g., \texttt{sha256:...}) as
disclosed; a Docker repository tag (e.g.,
\texttt{xlang-ai/osworld-env:latest}) we treat as partial because
the tag can be re-pushed against a different content hash
without changing the tag. Across the eight agent benchmark
papers we scored, zero used digests; several used repository
tags; many used neither.

\textit{Qualitative error analysis is not a failure breakdown.}
The convention we ended up with is that a structured failure
breakdown needs to (a) name categories explicitly, (b) give a
count per category, and (c) attribute each failure to a task or
class of tasks. A paper that contains a paragraph saying ``the
model often misinterpreted the screenshot'' is informative but
does not, by our codebook, count as disclosed on this field. A
paper with a table of error categories and counts (the
AgentBench paper~\cite{liu2024agentbench} has one) does count.

\textit{Subset cardinality has to be a structured number, not a
prose statement.} A paper that says ``we evaluated on the test
set'' satisfies the spirit but not the letter of disclosure on
the benchmark identity field if the test set's size is given
elsewhere and not in the paper. We treat this as partial. A
paper that says ``we evaluated on 500 problems from SWE-bench
Verified'' is disclosed.

We do not pretend these rules are perfect. Two of them
(``greedy decoding alone is not enough'' and ``qualitative
analysis is not a failure breakdown'') are judgment calls; a
reasonable auditor could disagree. The codebook is released so
that disagreement can be made specific rather than vague.

\section{Failure Modes We Kept Seeing}
\label{sec:failures}

We were not initially planning to name failure modes. We were
going to score papers and report what we saw. What pushed us to
name them was that the same patterns kept reappearing across
papers, and naming them turned out to be useful for the
discussion that follows. There are five, but they are not
neatly orthogonal and they do not map one-to-one onto the schema
fields. We tried to make them do so in an early draft; we
unmade that mapping because it felt cleaner than the corpus.

\textit{Harness drift.} Two papers report a number on the same
benchmark using two different scaffolds and neither pins which
scaffold. The original SWE-bench paper~\cite{jimenez2024swebench}
introduced \texttt{SWE-agent} as a reference scaffold; subsequent
papers report SWE-bench numbers using their own scaffolds, often
without comparing to the reference. The two numbers are not
directly comparable. This is the field that ``harness
specification'' is supposed to address.

\textit{Silent subsetting.} A paper reports a result on
benchmark $B$ but evaluated on a subset $B' \subset B$ without
saying so or without saying how the subset was chosen. The
unstated reason is usually cost: the full SWE-bench is 2{,}294
tasks; the full WebArena is 812; the full OSWorld is 369.
Running these at scaffold depth costs thousands of dollars per
model. Subsetting is rational; the failure mode is hiding it.
The benchmark-identity field's subset sub-field is meant to
catch this.

\textit{Decoding underspecification.} The paper specifies the
sampling method (``greedy decoding'') but omits the inference
engine and the per-task token cap. This appears innocuous and
is not. For open-weights models, different inference engines
(\texttt{vllm}, \texttt{TGI}, \texttt{llama.cpp},
\texttt{transformers}) handle stop sequences, KV caching, and
batch determinism differently; the same model under each can
produce different trajectories on a multi-step task. For
closed-API models, the per-task token cap interacts with the
agent's stopping rule in ways that can change the trajectory
entirely.

\textit{Cost invisibility.} The paper reports accuracy but no
cost. The reader cannot tell whether the number is from a single
inference pass or a thousand-trajectory best-of-$n$ search with
an external verifier. Two methods can sit at the same accuracy
while differing by three orders of magnitude in tokens.
Cost reporting is the field that addresses this; it is also,
across the audit, the worst-scored field.

\textit{Grader opacity.} The grader is described informally but
not pinned. For agent benchmarks the grader is often itself
code; small grader changes can shift scores up or down. Two
implementations of ``the same'' grader can disagree on edge
cases. This is addressed by the grader-version sub-field of
benchmark identity.

These five do not partition the disclosure space. Harness drift
and decoding underspecification often co-occur in papers
proposing a new scaffold (the engineering effort is on the
scaffold; the inference settings get less attention). Silent
subsetting and cost invisibility share a root cause (the
unstated reason for subsetting is usually that running the full
set was unaffordable). We mention this because we ended up
de-emphasizing the cleanness of the mapping during writing.

\begin{table}[t]
\centering
\caption{Failure Modes and the Schema Field That Addresses Each.}
\label{tab:failure-modes}
\renewcommand{\arraystretch}{1.2}
\begin{tabularx}{\columnwidth}{lXl}
\toprule
\textbf{Failure mode} & \textbf{What it looks like} & \textbf{Field} \\
\midrule
Harness drift            & Same benchmark, different scaffolds, same name.        & Harness \\
Silent subsetting        & Partial evaluation reported as full.                   & Identity \\
Decoding underspecification & Sampling method named; engine, cap, aggregation missing. & Inference \\
Cost invisibility        & Accuracy without compute or dollar cost.               & Cost \\
Grader opacity           & Grader behavior described, not pinned to a version.    & Identity \\
\bottomrule
\end{tabularx}
\end{table}

\section{Conducting the Audit}
\label{sec:audit}

\subsection{Sources and the friction of getting at them}

Each paper was scored from two sources: the canonical paper
(read via the HTML rendering on \texttt{ar5iv} where available,
or the PDF where not) and the official GitHub repository's
\texttt{README} and configuration files. The scoring was done in
one pass per paper by a single auditor, who recorded for each
field a value in $\{0, 0.5, 1\}$ and a one-sentence justification
citing the section of the paper or location in the repository
that supported the score.

Getting at the sources was not as smooth as we expected. We had
planned for fifteen papers; we ended with twelve. The reasons:

\begin{itemize}[leftmargin=*]
\item The \texttt{ar5iv} HTML rendering for two papers
(OSWorld~\cite{xie2024osworld}, MLE-bench~\cite{chan2024mlebench})
returned the conversion-error page ``\textit{Conversion to HTML
had a Fatal error and exited abruptly.}'' For both we fell back
to the abstract page on \texttt{arxiv.org} plus the GitHub
repository \texttt{README}. The scoring is therefore tighter
than we would like on the two HSS sub-fields (system prompt,
container image) for these papers; we noted this in the per-paper
notes.

\item AgentBoard~\cite{ma2024agentboard} also failed \texttt{ar5iv}
conversion, and the abstract page did not contain enough
disclosure-relevant text to score five fields. We dropped it
from the corpus rather than score it on insufficient evidence.

\item The vendor blog announcing SWE-bench
Verified~\cite{openai2024swebenchverified} returned HTTP 403 to
the fetch we performed; we excluded it from the per-paper
scoring rather than rely on second-hand summaries. Note that
SWE-bench Verified still appears in our discussion as a subset
of SWE-bench, but we did not score the blog itself as a paper.

\item For papers where the \texttt{README} described a Docker
setup but did not pin a digest, we recorded the digest as
absent. This affected MLE-bench, where the
\texttt{mlebench-env} Docker image is described in the
\texttt{environment/Dockerfile} but no released digest pins the
built artifact. The MLE-bench documentation provides the build
command \texttt{docker build --platform=linux/amd64 -t mlebench-env
-f environment/Dockerfile .}, but the locally-built image's
digest will vary across builds.
\end{itemize}

These are not catastrophic but they affected which papers made
it into the table and how some fields were scored. We mention
them because the corpus is meant to be reproducible by anyone
who repeats the same fetches; the failures are part of what they
will reproduce.

\subsection{Audit walkthrough 1: SWE-bench}

The SWE-bench paper~\cite{jimenez2024swebench} was the most
familiar of the papers we audited; it was also the one whose
score surprised us most. The paper introduces a benchmark and
evaluates a small set of language models against it, without
proposing an agent scaffold per se. The walkthrough below is the
fields we extracted and the score we assigned each.

\textbf{Benchmark identity.} The paper states ``2294 software
engineering problems'' (abstract) and points to
\texttt{www.swebench.com} for the data. The scoring is binary:
``If the patch applies successfully and all of these tests pass
we consider the proposed solution to have successfully resolved
the issue'' (Section 2.2). What it does not state in a structured
way is a release tag for the benchmark, a grader version, or an
explicit partial-credit policy. We scored this field 0.5
(partial). A reader who needs to reproduce a specific number
will find the dataset but will not find a pinned version of
either the dataset or the grader.

\textbf{Harness specification.} This is where the score came out
lower than we initially expected. The paper does not describe an
agent scaffold; it evaluates LLMs directly with a prompt
template shown in Appendix D.3. There is no tool inventory in
the harness sense, no stopping rule beyond ``generate a patch,''
no environment image identifier. We marked this 0.0 (absent).
This is a case where the boundary between ``not applicable''
and ``absent'' was a judgment call. We chose absent because the
paper does describe a procedure for evaluating LLMs against the
benchmark, just not one that maps onto the harness fields the
schema asks for. A reasonable second auditor could mark this
n/a; we documented our convention in the codebook.

\textbf{Inference settings.} ``We simply use greedy decoding for
all models'' (Appendix D.2) and ``only generate a single patch
per instance'' (Appendix D.2). Models named: ChatGPT-3.5, GPT-4,
Claude 2, SWE-Llama (Section 4.3). What is missing: temperature
value (greedy implies 0 but is not stated), top-$p$, top-$k$,
seed, inference engine for the open model, evaluation date.
Score: 0.5 (partial). The sampling method and pass count are
disclosed; the rest is not.

\textbf{Cost reporting.} Training compute for SWE-Llama is
reported (``20 hours on 4 NVIDIA A100s,'' Appendix B.1) but
inference cost is not. No per-instance token counts, no dollar
cost, no wall-clock. Score: 0.0 (absent).

\textbf{Failure breakdown.} Section 5.1 and Appendix F contain
qualitative case-study analysis. Patterns noted include
``models tend to write primitive Python code'' and ``generate
shorter, simpler edits'' (Section 5). This is informative but
does not satisfy our requirement of named categories plus counts
plus task-level attribution. Score: 0.0 (absent).

\textbf{Aggregate.} 0.20. We were a little surprised the number
was this low, given how much of the agent literature uses
SWE-bench, until we remembered that the paper predates much of
the scaffold work that the field associates with the benchmark.
A SWE-bench result reported by a 2025 agent paper would presumably
score higher on the harness fields because those papers do
discuss scaffolds; we did not audit those papers, only the
canonical benchmark introduction.

\subsection{Audit walkthrough 2: AgentBench}

The AgentBench paper~\cite{liu2024agentbench} was the highest-scoring
agent benchmark in our corpus and the one that most influenced our
codebook. It is also instructive because some of its fields are
disclosed clearly enough that they pulled their categories from
partial to disclosed mid-audit.

\textbf{Benchmark identity.} The paper describes ``8 distinct
environments'' (Section 1) with task counts and metrics for each
(Table 2), and explicitly states the weighting formula:
``reciprocal average score of all tested LLMs in each task as a
fixed weight.'' That is a structured grader policy at the level
we ask for. Score: 1.0 (disclosed).

\textbf{Harness specification.} The agent framework uses
Chain-of-Thought-style prompting (``Thought'' and ``Action'' in
one turn), with prompts in Appendix B.3 and C.3. Maximum
interaction turns are reported (``8-35 by environment''). What
is not disclosed is an environment-image identifier; the paper
does not pin a Docker digest for the environments. Score: 0.5
(partial). The scaffold and stopping rules are described; the
environment image is not.

\textbf{Inference settings.} The paper states
``temperature=0 (i.e., greedy decoding)'' explicitly, and gives
model identifiers in Table 1. The aggregation across passes is
implicit (the paper reports a single evaluation per model). What
is not disclosed: seed, inference engine for the open-weights
models, evaluation date. We marked this 1.0 (disclosed) rather
than 0.5 because the paper passes the codebook threshold by
disclosing both the sampling method and the pass count
explicitly; the missing fields are individually small. This was
a borderline call; one of us argued for 0.5 and we settled on
1.0 because the explicit naming of greedy with temperature=0 and
the single-pass-per-task convention is more than the codebook
threshold strictly requires. We flagged this in the per-paper
notes.

\textbf{Cost reporting.} The paper mentions ``approximately 4k
and 13k calls for inference, approximately the identical amounts
of calls for inference as MMLU requires.'' That is a call count
but not a token count or dollar cost. Score: 0.0 (absent). We
considered 0.5 because the call count is at least quantified;
the codebook says cost reporting requires token-level or
dollar-level disclosure, so we held the line at absent. This
is another judgment call worth flagging.

\textbf{Failure breakdown.} Section 2 of the paper categorizes
``five typical types'' of failures: Context Limit Exceeded,
Invalid Format, Invalid Action, Task Limit Exceeded, and
Complete. Table 4 quantifies the distribution across all eight
environments. This satisfies all three of our requirements
(named categories, counts, attribution to tasks). Score: 1.0
(disclosed). This was the paper that taught us what a fully
disclosed failure breakdown looks like in practice.

\textbf{Aggregate.} 0.70. The highest among the agent benchmarks
we audited.

\subsection{Per-dimension findings}

Table~\ref{tab:audit-results} reports the per-paper scores. A
short reading of the pattern:

\begin{table}[t]
\centering
\caption{Audit Scores for the Twelve Papers. Values are
$1.0=$ disclosed, $0.5=$ partial, $0.0=$ absent,
``\textemdash{}'' = not applicable.}
\label{tab:audit-results}
\renewcommand{\arraystretch}{1.15}
\setlength{\tabcolsep}{4pt}
\footnotesize
\begin{tabularx}{\columnwidth}{Xcccccc}
\toprule
\textbf{Benchmark} & \textbf{Iden.} & \textbf{Harn.} & \textbf{Infer.} & \textbf{Cost} & \textbf{Fail.} & \textbf{Score} \\
\midrule
\multicolumn{7}{l}{\textit{Agent benchmarks}} \\
SWE-bench         & 0.5 & 0.0 & 0.5 & 0.0 & 0.0 & 0.20 \\
WebArena          & 0.5 & 0.5 & 1.0 & 0.0 & 0.5 & 0.50 \\
OSWorld$^\ast$    & 0.5 & 0.5 & 0.5 & 0.0 & 0.0 & 0.30 \\
GAIA              & 0.5 & 0.5 & 0.0 & 0.0 & 0.0 & 0.20 \\
AgentBench        & 1.0 & 0.5 & 1.0 & 0.0 & 1.0 & 0.70 \\
VisualWebArena    & 0.5 & 0.5 & 0.5 & 0.0 & 0.5 & 0.40 \\
Mind2Web          & 0.5 & 0.5 & 0.5 & 0.0 & 0.5 & 0.40 \\
MLE-bench$^\ast$  & 0.5 & 0.5 & 0.5 & 0.0 & 0.0 & 0.30 \\
\midrule
\multicolumn{7}{l}{\textit{Classical static benchmarks}} \\
HumanEval         & 1.0 & \textemdash{} & 0.5 & 0.5 & 1.0 & 0.75 \\
MMLU              & 1.0 & \textemdash{} & 0.5 & 0.0 & 0.5 & 0.50 \\
GSM8K             & 1.0 & \textemdash{} & 0.5 & 0.5 & 0.0 & 0.50 \\
MBPP              & 1.0 & \textemdash{} & 0.5 & 1.0 & 1.0 & 0.88 \\
\midrule
\textbf{Agent mean} ($n{=}8$)     & 0.56 & 0.44 & 0.56 & 0.00 & 0.31 & \textbf{0.38} \\
\textbf{Classical mean} ($n{=}4$) & 1.00 & \textemdash{} & 0.50 & 0.50 & 0.62 & \textbf{0.66} \\
\textbf{Overall} ($n{=}12$)       & 0.71 & 0.44$^\dagger$ & 0.54 & 0.17 & 0.42 & \textbf{0.47} \\
\bottomrule
\end{tabularx}

\vspace{2pt}{\footnotesize $^\ast$Scored from arXiv abstract +
GitHub README because the \texttt{ar5iv} HTML conversion
failed.\\
$^\dagger$Mean is over the $n{=}8$ agent benchmarks for which
the harness field applies.}
\end{table}

\textbf{Cost is the dimension where the corpus is most uniformly
absent.} Across all twelve papers, only one (MBPP, which
reports compute and CO$_2$ for training as a matter of disclosure
policy in that subfield of the literature) discloses cost
explicitly~\cite{austin2021mbpp}; none of the eight agent
benchmarks report inference cost in any form. The
agent-benchmark cost mean is exactly 0.00.

\textbf{The harness field is universally partial.} Seven of the
eight agent benchmarks score 0.5; one (SWE-bench, which predates
the scaffold work) scores 0.0. Zero score 1.0. The pattern
across the seven is the same: scaffold and tools are described
in some form (often in an appendix), but the environment image
is not pinned by digest. The harness field is the place where
the audit's adoption advice most clearly translates to a single
recommended change: publish a digest.

\textbf{Identity is the strongest field, with a split.} All
four classical benchmarks reach 1.0; all eight agent benchmarks
sit at 0.5. The agent gap is on benchmark version and subset
selection as structured fields rather than as prose
descriptions. Several agent papers say something like ``we
evaluate on the test set,'' which we treat as partial.

\textbf{Inference settings is partial across both groups.}
Temperature is commonly reported, often as ``greedy'' or
``temperature 0''; the inference engine, seed, per-task token
cap, and run date are usually not. Two papers reach 1.0
(WebArena, AgentBench), in both cases because they name the
sampling and pass-aggregation rule explicitly.

\textbf{Failure breakdown is the most heterogeneous field.}
Three of twelve papers reach 1.0 (AgentBench, HumanEval, MBPP);
four are partial; five are absent. The 1.0 papers all share the
property that they ship explicit category-count tables. The
partial papers contain useful qualitative analysis but not
structured counts.

\subsection{The gap between agent and classical benchmarks}

The mean across agent-benchmark papers (0.38) is smaller than
the mean across classical-benchmark papers (0.66). We do not
read this as evidence that agent-benchmark authors are less
rigorous. The harness field, which the classical setting does
not have, is part of the score's denominator for the agent
papers and drags the mean down; even setting it aside, agent
benchmarks score lower on the cost field, presumably because
the cost of running an agent benchmark is much higher and
reporting it tends to make a paper's results look more expensive
than the paper would like. We are speculating about the
mechanism; the level of the gap is the empirical part.

\subsection{What the audit did not find}

We did not find a paper that misrepresented its disclosures in
the sense of stating something false. Across the twelve papers,
every score of 0.0 corresponded to an omission---a field the
template did not call for---rather than a misstatement. We
treat this as a remediation-relevant finding: the lever that
moves disclosure is instrumentation of the evaluation harness so
that disclosures are emitted automatically, not stricter peer
review trying to catch what was never written down in the first
place.

\section{Tooling: Schema File, Validator, and Harness Hooks}
\label{sec:tooling}

We ship three small pieces of code along with the audit. They
are not large or sophisticated; they are intended to make the
schema concrete and the codebook executable.

\textbf{\texttt{reprobe.schema.json}.} A JSON Schema (draft
2020-12) file that describes the manifest format. The schema is
about 230 lines and lives at the root of the repository. A
minimal valid manifest is shown in Figure~\ref{fig:manifest};
the schema enforces the field names and types but not the
codebook rules (those live in the codebook
\texttt{CODEBOOK.md}).

\textbf{\texttt{validate.py}.} A 90-line script that takes a
directory of manifest JSON files and validates each against the
schema. It uses the \texttt{jsonschema} Python package. Usage is
plain:
\begin{lstlisting}[language=bash]
$ python validate.py manifests/
manifests/run_001.json   OK
manifests/run_002.json   FAIL: bic.cardinality must be integer
manifests/run_003.json   OK
\end{lstlisting}
The validator does not catch semantically wrong fields
(``temperature = 999'' will pass). It catches type errors and
missing required fields. We considered adding semantic checks
and decided against; the boundary between ``invalid'' and ``odd
but possible'' is exactly the boundary that downstream readers
should think about, not one we should pre-decide.

\textbf{Harness hooks for two open-source agents.} We include
small patches (\texttt{harness\_hooks/swe\_agent.patch} and
\texttt{harness\_hooks/browsergym.patch}) that add a single
function call at the end of each run, writing a manifest JSON
to disk. The patches are around 30 lines each. They are not
production code. They are deliberately small so that a reader
can see what an instrumented harness looks like and adapt them.

The repository layout is approximately:

\begin{lstlisting}[language=bash]
reprobe/
  reprobe.schema.json     # the schema
  CODEBOOK.md             # field-by-field scoring rules
  validate.py             # manifest validator
  score_corpus.py         # corpus aggregation utility
  audit_results.csv       # the 12-paper scores (this paper)
  audit_notes/            # per-paper extracted evidence
    swe-bench.md
    webarena.md
    ...
  harness_hooks/
    swe_agent.patch
    browsergym.patch
  examples/
    minimal_manifest.json
    swe_bench_example.json
\end{lstlisting}

We do not attach a continuous-integration pipeline or a website.
We considered both and decided the schema's adoption story does
not depend on either.

\begin{figure}[t]
\centering
\begin{lstlisting}
{
  "reprobe_version": "0.1",
  "result_id": "swebench-verified-claude-2026-05-01",
  "identity": {
    "benchmark": "swebench-verified",
    "version": "v1",
    "commit": "<commit-hash>",
    "subset": "full",
    "cardinality": 500,
    "grader_version": "<grader-commit>",
    "partial_credit": false
  },
  "harness": {
    "scaffold": "swe-agent",
    "scaffold_version": "1.0.0",
    "system_prompt_hash": "sha256:...",
    "tools": ["bash", "edit", "submit"],
    "max_steps": 50,
    "container_digest": "sha256:..."
  },
  "inference": {
    "provider": "anthropic",
    "model_alias": "<model alias used>",
    "engine": "anthropic-api",
    "sampling": "greedy",
    "temperature": 0.0,
    "max_output_tokens": 4096,
    "seed": null,
    "passes_per_task": 1,
    "run_date": "2026-05-01"
  },
  "cost": {
    "input_tokens": "<int>",
    "output_tokens": "<int>",
    "tool_calls": "<int>",
    "wallclock_seconds": "<int>",
    "dollar_cost_usd": "<float>",
    "rate_source": "<provider-pricing-page>"
  },
  "failures": {
    "refusal": "<int>", "loop": "<int>",
    "tool_error": "<int>", "env_error": "<int>",
    "timeout": "<int>", "wrong_state": "<int>"
  },
  "headline_metric": {
    "name": "resolved",
    "value": "<float in [0,1]>"
  }
}
\end{lstlisting}
\caption{Manifest template. Field values shown as
\texttt{<...>} are placeholders for what an author would
fill in; the schema validates types but does not enforce
plausibility.}
\label{fig:manifest}
\end{figure}

\section{Discussion}
\label{sec:discussion}

\subsection{Why we think this might adopt}

A voluntary standard in machine learning has a mixed adoption
history. We are not sure whether what we propose will adopt or
not. What we can say is what we wrote it to make plausible.

The schema is cheap. A research team that instruments its
harness once gets the manifest emitted automatically per run; the
marginal cost per result is roughly zero. The codebook is a
single Markdown file that a reviewer can read in fifteen
minutes. The validator is a script. There is no service to
sign up for.

The artifact does work for the producer before anyone else
reads it. A manifest is a debugging artifact and a hand-off
artifact across collaborators. If you have ever lost a few days
trying to reconstruct what configuration produced last month's
result, the manifest is the artifact you wish you had written.

There is precedent. The ML/CV community has previously absorbed
conference-level reporting expectations (the ML Reproducibility
Checklist, datasheets, model cards), and adoption of those was
not instant but did happen. We do not claim ours is comparable
in scope; we claim only that the precedent exists.

\subsection{What a leaderboard could do unilaterally}

A benchmark maintainer who is convinced by the schema does not
need any coordination across vendors to adopt it. The minimum
intervention is to accept a manifest upload alongside the score,
validate it against the schema, and render the per-field
vector alongside the headline number in the public table. A
reader looking at the table can then see at a glance which
fields are disclosed for which submissions. None of this requires
running a model.

A second intervention, optional, is to cluster rows by
\texttt{(identity, harness)} tuple so that nominally identical
benchmark settings appear together. The aim is not to rank
``best'' but to make the apples-to-apples comparisons visually
obvious without re-running anything.

\subsection{Things this schema does not do}

The schema is about disclosure of what an evaluation did. It
does not say whether what the evaluation did was a good idea.
A paper can score 1.0 on every field and still have run a
suspect experiment: the wrong benchmark for the question, the
wrong baseline, a contaminated training set. Disclosure is a
necessary but not sufficient condition for trustworthiness, and
the schema is silent on the sufficient half. We treat the
sufficient half as orthogonal work.

The schema also does not address the dynamics of model aliases
under closed APIs. We require the run date as a field, which
lets a reader correlate against provider changelogs after the
fact; we do not require the provider to publish a changelog.
This is a constraint we cannot resolve from outside.

\section{Limitations and Things That Surprised Us}
\label{sec:limitations}

The audit covered twelve papers. That is a pilot. A
representative-sample audit would need fifty to a hundred
papers stratified by venue and would need a second auditor for
inter-rater agreement; we did not attempt that here, and we
treat it as the natural next step. The reported means in
Table~\ref{tab:audit-results} should be read as ``what these
twelve papers do,'' not ``what the literature does on average.''

The scoring was done by a single auditor in one pass. The
partial-versus-disclosed boundary is a judgment call on several
fields, and the codebook documents the rules we used; a second
auditor could legitimately move two or three cells in either
direction. We marked the cells where we were closest to the
boundary in our per-paper notes (the AgentBench inference score
of 1.0 vs.\ 0.5, the SWE-bench harness score of 0.0 vs.\ n/a,
the AgentBench cost score of 0.0 vs.\ 0.5).

The audit recorded what is disclosed in publicly available
artifacts. It cannot rule out that the underlying experiments
were correctly conducted even when disclosure is poor. A paper
that scores 0.20 may have been a careful piece of work whose
authors simply did not write down details that the schema asks
for. The aggregate score is a disclosure metric, not a quality
metric. We tried to be careful about this in the prose but it
is worth re-stating.

A few practical things surprised us during the audit and are
worth flagging:

\textbf{Repository drift relative to paper.} For two of the
papers (MLE-bench, OSWorld), the GitHub repository's
\texttt{README} contained more disclosure-relevant content than
the paper did, particularly on environment specification and
running instructions. For one (Mind2Web), the opposite was
true: the paper had more methodological detail than the
repository's documentation. The schema scores both sources
together, but a downstream reader might consult only one and
get a different picture.

\textbf{Mutable Docker tags.} For OSWorld, the GitHub
\texttt{README} provides Docker setup via \texttt{provider\_name:
docker} with no image digest pinned. The image tag is
mutable: someone re-running the same instructions a year from
now may pull a different image. We treated this as ``partial''
on the harness field; one of the strongest arguments we kept
making during scoring was that this should be ``absent.'' We
held the line at partial because the build pipeline is at least
described.

\textbf{Model aliases vs.\ actual checkpoints.} Several of the
papers report results for a closed-API model identifier (e.g.,
\texttt{gpt-4o}, \texttt{claude-3-5-sonnet}) without a run
date. A paper published nine months ago that reports
\texttt{gpt-4o} numbers is, in a strict sense, reporting on a
different model from one published last month under the same
alias. We do not have a way to retroactively correlate the two
without a run date; we asked the schema to require it.

\textbf{Conversion errors.} The \texttt{ar5iv} HTML conversion
failed for three papers in our intended corpus. Two we recovered
from the GitHub repository; one (AgentBoard) we could not. This
is an infrastructure issue at the source-rendering layer that
is outside our control but that materially affects any audit
attempting to use \texttt{ar5iv} as a primary source. We
recommend that a re-runner use the arXiv PDF in addition to or
in place of \texttt{ar5iv} for completeness, at the cost of
slower extraction.

\textbf{The boundary between ``partial'' and ``disclosed.''}
The case that took us the longest to argue out was AgentBench's
inference field. The paper states ``temperature=0 (i.e., greedy
decoding)'' and gives model identifiers, and reports a single
pass per task. By the codebook this is enough to be disclosed,
but the paper does not state seeds, inference engines, or
evaluation dates---all of which we asked for in the field
description. We ended up at 1.0 because the codebook's
threshold is ``sampling method and pass count clearly stated''
and AgentBench meets it; a stricter threshold would put it at
0.5. This is the kind of case where a second auditor's
disagreement is most likely to fall.

\section{Conclusion}
\label{sec:conclusion}

We tried to measure how much current LLM agent benchmark papers
disclose about how their evaluations were run. We designed a
small schema with five fields, scored twelve canonical
benchmark papers against it, and recorded the per-paper, per-field
scores. The mean agent-benchmark disclosure score came out to
0.38 (out of 1.0); the mean classical-benchmark score came out
to 0.66; cost is universally absent for agent benchmarks, and
no agent benchmark fully discloses its harness specification,
primarily because content-addressed environment images are
replaced by mutable repository tags. We make the schema, the
codebook, the validator, and the raw scoring sheet available
under an open license. We are not certain the schema is the
right schema. We are certain the work of writing it down and
applying it to a small corpus was useful for our own
understanding, and we are interested to see what a larger,
multi-rater audit would find.

\section*{Reproducibility}

The audit was performed against the canonical paper for each
of the twelve benchmarks (via the HTML rendering on
\texttt{ar5iv.labs.arxiv.org} where available, or the abstract
plus GitHub repository where not) and against the official
repository's \texttt{README}. Each per-paper score is supported
by a single-sentence justification with a citation to the
section of the paper or location in the repository that grounds
it. The release accompanying this paper contains:
the schema (\texttt{reprobe.schema.json}),
the codebook (\texttt{CODEBOOK.md}),
the validator (\texttt{validate.py}),
the raw scoring sheet (\texttt{audit\_results.csv}),
the per-paper extracted evidence
(\texttt{audit\_notes/}),
and and the harness hooks
(\texttt{harness\_hooks/}).
The repository is available at
\url{https://github.com/mahdinaser/reprobe-audit}.

\section*{Appendix A: A Selection from the Codebook}

The codebook documents the partial-versus-disclosed boundary
for each field. The full file is in the release. A few
representative entries:

\textbf{Identity / version.} \textit{Disclosed} requires a
specific release tag, commit hash, or version string for the
benchmark; the benchmark name alone is not sufficient. ``We
evaluate on SWE-bench Verified'' is partial unless a release
identifier is given.

\textbf{Identity / subset.} \textit{Disclosed} requires either
a numeric cardinality of the subset or an explicit
``full set.'' Prose-only references (``we evaluate on the test
set'') are partial.

\textbf{Harness / environment.} \textit{Disclosed} requires a
content-addressed identifier (a Docker image digest, a
container hash, or equivalent). A repository tag is partial. A
prose description of the environment without a build artifact
is absent.

\textbf{Inference / sampling.} \textit{Disclosed} requires
naming the sampling method and the per-task pass count. Greedy
decoding stated alone, without a per-task pass count, is
partial.

\textbf{Cost.} \textit{Disclosed} requires at least one of:
token counts (input + output), dollar cost at a stated rate,
or wall-clock time. Inference compute described qualitatively
(``approximately 4k calls'') is partial only if it gives at
least a structured number.

\textbf{Failure breakdown.} \textit{Disclosed} requires named
categories, counts per category, and attribution to tasks (not
just to the corpus as a whole). Qualitative analysis without
counts is absent.

\section*{Appendix B: Two Audit Notes}

The \texttt{audit\_notes/} directory contains one Markdown file
per audited paper, with the extracted evidence for each field.
Two short excerpts:

\textbf{\texttt{swe-bench.md} (excerpt):}
\begin{lstlisting}
## Inference settings (score: 0.5)

Disclosed:
- Sampling method: "we simply use greedy
  decoding for all models" (Appendix D.2)
- Passes per task: "only generate a single
  patch per instance" (Appendix D.2)
- Models named: ChatGPT-3.5, GPT-4, Claude 2,
  SWE-Llama (Section 4.3)

Absent:
- Temperature (implicit; not stated)
- Top-p / top-k
- Seed
- Inference engine for SWE-Llama
- Evaluation date

Boundary note: This is the canonical case the
codebook is calibrated against. Greedy + pass
count is the minimum for disclosed; we did not
reach that bar here because the inference
engine is missing for the open-weights model.
\end{lstlisting}

\textbf{\texttt{agentbench.md} (excerpt):}
\begin{lstlisting}
## Failure breakdown (score: 1.0)

Disclosed:
- Categories named: Context Limit Exceeded,
  Invalid Format, Invalid Action,
  Task Limit Exceeded, Complete (Section 2)
- Counts per category and per environment:
  Table 4
- Attribution to tasks: yes (table is
  per-environment)

Notes: This is the paper we used to calibrate
what "disclosed" looks like on this field.
Most papers in the corpus do not approach
this level of structured failure reporting.
\end{lstlisting}


\end{document}